\documentclass[11pt,a4paper]{article}
\usepackage[hyperref]{acl2019}
\usepackage{times}
\usepackage{latexsym}
\usepackage[utf8]{inputenc}
\usepackage{amsmath}
\usepackage{amssymb}
\usepackage{latexsym}
\usepackage{amsthm}
\usepackage{qtree}

\usepackage[english]{babel}
\usepackage{setspace}
\usepackage{braket}
\usepackage{mathtools}
\usepackage{graphicx}
\usepackage{linguex}
\usepackage{bussproofs}
\usepackage{lscape}
\usepackage{tikz}
\usepackage{graphicx}
\usepackage{leftidx}
\usepackage{tikz-cd}
\usepackage{hyperref}
\usepackage{listings}
\usepackage{float}
\usepackage[makeroom]{cancel}
\usepackage{xcolor}
\usepackage[inline]{enumitem}
\usepackage{bm}
\usepackage[noabbrev,capitalize]{cleveref}
\usepackage[titletoc,title]{appendix}

\aclfinalcopy

\crefformat{section}{\S#2#1#3} 
\crefformat{subsection}{\S#2#1#3}
\crefformat{subsubsection}{\S#2#1#3}

\allowdisplaybreaks

\tikzset{>=latex}

\newcommand{\norm}[1]{\left\lVert #1 \right\rVert}

\usepackage[autostyle, english = american]{csquotes}
\MakeOuterQuote{"}

\begin{document}
\title{Distributed neural encoding of binding to thematic roles}

\author{Matthias Lalisse\\
Dept of Cognitive Science\\
Johns Hopkins University\\
Baltimore, MD USA \\
{\tt lalisse@jhu.edu} \And
Paul Smolensky\\
Dept of Cognitive Science\\
Johns Hopkins University\\
\& Microsoft Research AI \\
Redmond, WA USA\\
{\tt smolensky@jhu.edu}
}

\maketitle

\date{}

\maketitle

\begin{abstract}
A framework and method are proposed for the study of constituent composition in fMRI. The method produces estimates of neural patterns encoding complex linguistic structures, under the assumption that the contributions of individual constituents are additive. 
Like usual techniques for modeling compositional structure in fMRI, the proposed method employs pattern superposition to synthesize complex structures from their parts. Unlike these techniques, superpositions are sensitive to the structural positions of constituents, making them irreducible to structure-indiscriminate ("bag-of-words") models of composition. Using data from a study by \citet{frankland_greene_2015}, it is shown that comparison of neural predictive models with differing specifications can illuminate aspects of neural representational contents that are not apparent when composition is not modelled. The results indicate that the neural instantiations of the binding of fillers to thematic roles in a sentence are non-orthogonal, and therefore spatially overlapping.
\end{abstract}


Casual inspection of human linguistic behavior reveals a striking fact: the very same words, in very slightly different configurations, correspond to very different situations and events. While it is simple to state this fact in a symbolic theory of cognition, the question naturally arises: how is this structure-sensitivity of linguistic representation|ultimately the product of neural processes|realized in the kinds of representational resources available to the brain? 
We will refer to this as the \textit{Structure Encoding Problem}. 
The present work investigates the human brain's solution to this problem for the encoding of propositions, where the distinct structural assignments of \textit{cat} and \textit{dog} in the propositions expressed by "the cat chased the dog" and "the dog chased the cat" crucially distinguish their meanings. Analyzing relevant fMRI neuroimaging data generated by the groundbreaking work of \citet{frankland_greene_2015}|henceforth F\&{}G|we contrast the simple, localist solution to the Structure Encoding Problem considered by F\&G with a distributed hypothesis. This distributed hypothesis is derived from theoretical work in AI on the artificial-neural-network version of this problem \citep{smolensky1990tensor}, where an explicit account of distributed encoding of symbolic structure|the Tensor Product Representation|is developed, unifying many previous approaches. The new analysis supports the distributed hypothesis.\footnote{Presented as a poster at \href{https://www.macsim.us/macsim-8-nyu/}{MACSIM 8} (2019).}

\cref{sec_problem_statement} elaborates on the neural version of the Structure Encoding Problem, putting forward two hypotheses. 
The overall approach pursued in the new analysis of F\&{}G is summarized in \cref{sec_reanalysis}. Our method for arbitrating between the hypotheses is put forth in \cref{sec_models}, and the results are presented in \cref{sec_results}. Discussion and conclusions follow in \cref{sec_discussion} and \cref{sec_conclusion}. Two appendices provide technical details.

\section{Problem statement} \label{sec_problem_statement}

Following F\&G, we will refer to the structural role filled by \textit{(the) cat} in the proposition expressed by "the cat chased the dog" as the \textit{agent} and will use $\text{cat}_a$ to denote the \textit{binding} of \textit{cat} to the role \emph{agent}. The element that fills a role such as agent will be referred to as its \textit{filler}, leaving open whether this element is lexical, syntactic, conceptual, etc. In the case of a proposition, the fillers are semantic or conceptual. 

The proposition expressed by "the cat chased the dog" is assumed to be encoded by combining the filler-role binding $\text{cat}_a$ with the filler-role binding $\text{dog}_p$, which respectively denote the binding of "cat" to the agent role and "dog" to the patient role. This proposition is denoted $\langle \text{cat}_a, \text{dog}_p \rangle$. Below, we will not be concerned with the encoding of the verb, so it is omitted from the notation.

The following hypotheses are here entertained. \break \break
\textbf{Hypotheses}

\ex. \textbf{General version.} \label{hypothesis_general}
The neural encoding of the proposition $\langle \text{cat}_a, \text{dog}_p \rangle$ is the vector sum (superposition) of the activation patterns encoding $\text{cat}_a$ and $\text{dog}_p$: $\text{cat}_a + \text{dog}_p$

\ex. \textbf{Localist version.} \label{hypothesis_local}
The activation patterns for agent bindings $\text{X}_a$ and patient bindings $\text{Y}_p$ reside on disjoint sets of units (which entails that ${\text{X}_a}^\top \text{Y}_p = 0$).

\ex. \textbf{Distributed version.} \label{hypothesis_distributed}
The units supporting the activation patterns for agent bindings $\text{X}_a$ and patient bindings $\text{Y}_p$ are not disjoint (under which it is possible that ${\text{X}_a}^\top \text{Y}_p \neq 0$).

The content of \ref{hypothesis_general} is that filler-role bindings are associated with points in the neural state space, which are combined via pattern superposition. This constitutes an explicit proposal for the neural realization of compositionality, i.e. of how complex structures are built from simpler neural elements. The content of \ref{hypothesis_local} is that the representation spaces for each role are spatially disjoint, and the content of \ref{hypothesis_distributed} is that they are not.

For present purposes, the vectors cat$_a$ and dog$_p$ are treated as primitive. The related but distinct question of how these patterns might be built systematically|perhaps using operations applied to structure-indiscriminate fillers and mapping these to their role-bound instantiations|is obviously fundamental. Several detailed theoretical proposals exist to this effect \citep{smolensky1990tensor, plate1994thesis, kanerva2009hyperdimensional}, though it is not simple to relate them to available neural observations. 

The Structure Encoding Problem calls for an explanation of the fact that "The cat chased the dog" is cognitively distinct from "The dog chased the cat". It is not obvious that at the neural level these distinctions would appear as some kind of combination of neural patterns representing each constituent of the proposition|i.e. of filler-role bindings|as is suggested by the idea of compositionality interpreted very literally (as in \ref{hypothesis_general}).\footnote{In applied work, \citet{ettinger2018composition} searched for information about the agents of sentences in various engineered sentence representations, using a multilayer perceptron with a nonlinear activation function (ReLU). What distinguishes such classifiers from the linear classification methods standard in cognitive neuroscience is that, in linear methods, each class (filler-role binding) is associated with a point or contiguous set of points in the brain's representational space. As in the XOR problem, solutions attained by nonlinear classifiers cannot generally be interpreted in this manner. For instance, the mean of the TRUE cases in the XOR problem is not a TRUE case. 
}
It is certainly reasonable to adopt something of the sort as a first hypothesis, and this is reflected in the fact that the leading theory of generative grammar with a well-articulated connectionist foundation|Optimality Theory \citep{prince1997optimality}|assumes the existence of such representations, and provides the technical means for instantiating them as activity patterns over neural units. 

Moreover, several experimental results suggest that this literal interpretation is in fact correct. In a widely-known study, data from which are reanalyzed in \cref{sec_reanalysis}-\ref{sec_discussion}, \citet{frankland_greene_2015} found a pair of brain regions in the superior temporal sulcus, adjacent but non-overlapping in their searchlight analysis, that selectively carried information about the identities of agents and patients in sentences like "The cat$_{agent}$ chased the dog$_{patient}$." In the agent region, decoding was significant for agents but not patients, and symmetrically for the patient region. This finding is compatible with Claim \ref{hypothesis_general}, but since the regions supporting decoding of each role are spatially disjoint (as in \ref{hypothesis_local}), this is superposition of a somewhat uninteresting kind. 
\begin{quote}
At a high level, these regions may be thought of as functioning like the data registers of a computer, in which time-varying activity patterns
temporarily represent the current values of these variables. This functional architecture could support the compositional encoding of sentence meaning involving an agent and a patient, as these representations can be simultaneously instantiated in adjacent regions to form complex representations with explicit, constituent structure. \citep{frankland_greene_2015}
\end{quote}
The presence of distinct neural patterns encoding distinct filler-role bindings may likewise be  inferred from the results of \citet{wang2016thematicroles}.
\footnote{Though using stimuli in the visual modality and without an explicit proposal for the representational architecture.} 


All of this implies that, with respect to the filler-role binding problem, models of neural composition that are based on summation of structure-indiscriminate pattern components \citep{anderson2016LatentWords, pereira2018decoding} are clear non-starters.\footnote{An apparent exception is work by A.J. Anderson et. al., who employ an additive model of structure-indiscriminate word-level features combined into sentence-level representations by averaging. These sentence models are regressed against fMRI observations of subjects reading the corresponding sentences, modeling the neural patterns associated with entire sentences. The result of this sentence-model-to-brain mapping is then pseudo-inverted to produce a brain-to-model decoder. By selectively removing the contribution to the sentence model associated with words in specific grammatical positions|e.g. removing the direct object "the powerful hurricane" from the sentence "The family survived the powerful hurricane", the authors asked whether decoding accuracy decreased in each region. The alternative hypothesis targeted by this approach is that there are brain regions that exclusively contain information about specific lexical or phrasal classes. Such exclusivity would be revealed by a region's failure to exhibit decreased decoding accuracy when information from the other classes is omitted \citep{anderson2018GrammaticalPosition}. As applied there, the resolution of this approach|both spatially and theoretically|is limited.} The necessary distinctions between patterns for structures as simple as "mountain gold" and "gold mountain" are unavailable to such approaches even in principle \citep{baron2011composition}.\footnote{Note that, while vector addition itself is commutative, combination of filler-role bindings is not, because the pattern components cat$_a$ and and cat$_p$ are, by hypothesis, distinct (see \ref{def_role_discr}).} 

\section{Reanalysis} \label{sec_reanalysis}

This section describes an exploratory reanalysis of data from Experiment 2 of \citep{frankland_greene_2015} shared with the authors by Steven Frankland. In line with Claim \ref{hypothesis_general} that the contributions of individual filler-role bindings to the representation of a proposition are additive, a linear "forward" model of patterns encoding full propositions was iteratively fit to a subset of the experimental trials. The model thus fit was then evaluated on a held-out set of data, using a predict-to-decode methodology that is conventional in this domain \cite{mitchell2008decoding}. 

Our central manipulation involves a model comparison. As mentioned in \cref{sec_problem_statement}, the results reported in F\&{}G are compatible with a superposition account of filler-role binding in a rather uninteresting sense, equivalently expressed as a kind of vector concatenation. 
However, a decoding methodology based on independent, single-role decoding is unable to arbitrate decisively between hypotheses \ref{hypothesis_local} and \ref{hypothesis_distributed}. The data from each trial of Experiment 2 are "mixed", meaning that a trial with $\langle \text{man}_a, \text{cat}_p\rangle$ contains pattern components from both man$_a$ and cat$_p$, and similarly for a trial with $\langle \text{man}_a, \text{girl}_p\rangle$. In an ROI that contains information only about agents, the patterns cat$_p$ and girl$_p$ will, on average, be identical. If the ROI is, in fact, sensitive to patients, then these patterns will vary systematically as a function of the patient in the trial. From the point of view of a model of agents where the patients are ignored, this systematic variation will appear to be noise. However, if the signal associated with patients in this ROI is small relative to the signal associated with agents, this additional unmodelled variation may not significantly disrupt decoding. Turning things around, consider the patient patterns when the patient signal is weaker than the agent signal. Though this signal may be present in the region, it may be small relative to the variation associated with the unmodelled agent, leading to a failure to decode the patients. This last case would represent a false negative. 

A stronger test of orthogonal representation spaces, then, is provided by explicitly comparing models that do and do not model pattern composition. To this end, we evaluate two classes of predictive models. In the \emph{single-pattern} models, patterns for the fillers in the agent and patient roles are estimated in independent linear regressions, and are also independently compared with each held-out image in decoding. Our central manipulation is to fit \emph{mixed-pattern} models that estimate filler patterns for both roles within each region. Then, when decoding experimental conditions from held-out trials, patterns are synthesized from the learned regression coefficients for \emph{both} roles, modeling the entire proposition, rather than just one of its constituents. Hence, the mixed model predicts, in addition to the pattern for the role being decoded, the value of the other role in the given proposition. Specifically, each proposition type is modeled as a superposition of pattern components for both filler-role bindings. 

If a given region provides information about the contents of one role, but not the other, then explicitly modeling the contents of the role that is not represented will have no effect on decoding accuracy, and may even decrease it. On the other hand, if informative patterns for both roles are superposed within a single ROI, then inclusion of information about the contents of the other role will improve the match between predictions and actual patterns. Under orthogonal representation,then,  we predict that there should be no advantage to decoding with this mixed-pattern model. If, however, the inclusion of information about the other role does affect decoding, we conclude that a region contains information about both roles. These predictions are derived in greater detail in Appendix B. 

\subsection{Data}

Frankland and Green's 25 subjects underwent fMRI while reading transitive sentences constructed by crossing four nouns (man, girl, dog, cat) with five verbs (chased, scratched, bumped, approached, blocked), omitting the diagonal of the nouns (e.g. "The cat chased the cat"). Sentences occurred an equal number of times in active and passive syntactic configurations, which are treated as identical in the data coding ("The cat chased the dog" = "The dog was chased by the cat"). Each trial consisted of a 3.5-second sentence presentation, followed by 7.5 seconds of fixation, and then a comprehension question on 1/3 of trials. The $4\times 5\times 3 = 60$ unique propositions were seen 6 times across 6 runs of the experiment, yielding 360 experimental trials. fMRI images with 1.5-mm$^3$ isotropic voxels were obtained at an interval of 2.5 seconds and spatially smoothed with a 1.5mm$^3$ full-width half-maximum Gaussian kernel. To produce a single image for each trial, images from the 7.5-second interval following sentence presentation were averaged across time.  

We consider data from two regions of interest (ROIs) that were found to be agent-selective (ROI-A) and patient-selective (ROI-P) in F\&{}G's analysis. These regions were localized for post-hoc analyses using a leave-out-one-subject localization procedure detailed in the Supporting Information of \citep{frankland_greene_2015}. 

\section{Modeling filler-role bindings} \label{sec_models}

\begin{figure*}[h!]\centering
\includegraphics[scale=.2]{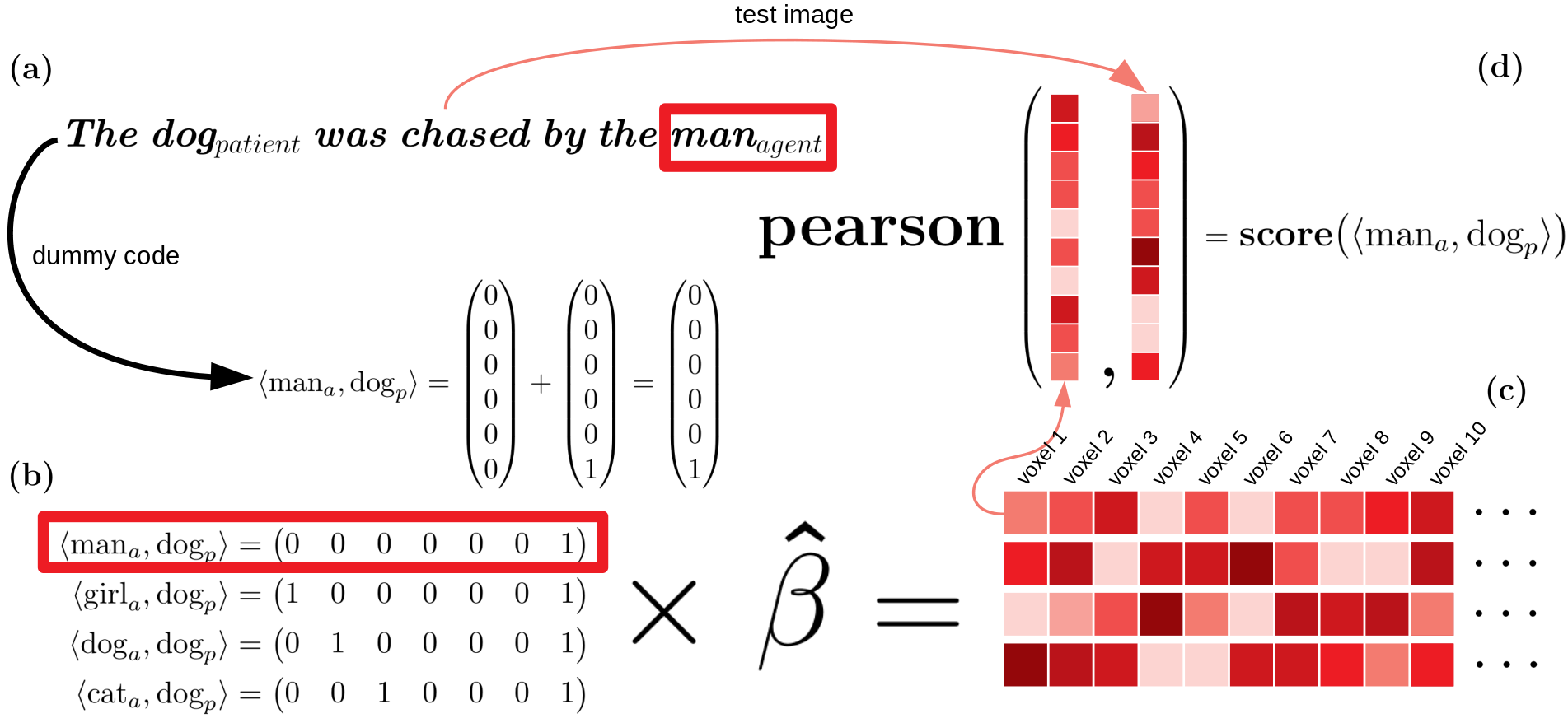}
\caption{\textbf{Mixed-pattern decoding}. To decode the agent of the sentence \emph{The dog was chased by the man} in a held-out trial, (a) indicator codes for the four possible propositions with \emph{dog} as patient are produced. Each indicator-coded proposition is the sum of an indicator code for the agent argument, and an indicator code for the patient argument. (b) The indicator-coded propositions are then multiplied by the pattern estimates $\hat{\beta}$ to produce (c) predicted patterns for each proposition. (d) Each prediction is then compared with the image of the held-out trial using the Pearson correlation. The best-correlated predicted pattern is chosen as the most likely label for the test instance. 
 }\label{fig_decoding}
\end{figure*}

Using a training set of experimental trials, the models were fit by linear regression (Generalized Least Squares) to produce pattern estimates for each regressor. Three linear model specifications were considered: one including only agents as regressors, another including only patients, and the mixed-pattern specification modeling two-pattern superpositions. Model estimation in each case yielded a matrix of regression coefficients $\hat{\beta}$ whose columns are predicted values for each voxel in the ROI, corresponding to each regressor (rows). 

\subsection{Two technical remarks} \label{sec_technical_remarks}

\textbf{Multicollinearity}. From the point of view of pattern estimation in an experiment like F\&{}G's, a difficulty arises which is more or less inescapable within this problem domain. The difficulty is that, in any multiconstituent structure, multiple modelable pattern components appear in each observation, making the data "mixed". 
In many cases, including the present one, this leads to multicollinearity in the design matrix of the forward model when all constituents are modelled, corresponding to the fact that the contributions of individual constituents to the mixed pattern cannot be uniquely decomposed. We resolve this estimation problem by employing an indicator code for each filler-role binding as detailed in Appendix A. This yields a predictive model of the mixed data|i.e. a prediction for each propositional pattern|but the estimated coefficients are not interpretable as patterns for individual filler-role bindings. 

\noindent
\textbf{Multivariate noise whitening}. Data from fMRI are very noisy, and the noise across measurement channels (voxels) is highly correlated. This means 
\begin{enumerate*}[label=Case \arabic{enumi}.,ref=Step \arabic{enumi}]
\item[(a)] that particularly noisy voxels are more likely to deviate from their expected values under the experimental conditions, and 
\item[(b)] that the degree of deviation from this value covaries across voxels.
\end{enumerate*}
To correct for this, the data were whitened by multivariate noise-normalization \citep{diedrichsen&kriegeskorte2017representational_models} using a regularized estimate of the noise covariance matrix obtained from just the training data in each fold (Appendix A). Univariate noise-normalization (division of a voxel's value by the voxel's standard deviation across measurements, ignoring all other voxels) was also experimented with, but all models performed systematically worse under that regimen. 

\subsection{Decoding with single- and mixed-pattern models} \label{sec_decoding_description}

A set of neural patterns deserving the name "filler-role binding" should exhibit two properties:\\\vspace{.2cm}
\textbf{Properties of a filler-role binding}
\vspace{-.2cm}
\ex. \textbf{Role-discriminate}. cat$_a$ and cat$_p$ are distinct. \label{def_role_discr}

\vspace{-.7cm}
\ex. \textbf{Filler-discriminate}. cat$_a$ and dog$_a$ are distinct.\label{def_filler_discr}

\ref{def_role_discr} asserts that the representations are sensitive to the assignment of the filler to the role. \ref{def_filler_discr} requires that these patterns additionally discriminate between each filler within a given role. Although analytic measures of pattern distinctness exist \citep{allefeld&haynes2014manova,kriegeskorte2007faces}, the traditional way of verifying pattern distinctions like \ref{def_role_discr} and \ref{def_filler_discr} is to attempt to decode cognitively distinct variables from neural observations, which is the strategy adopted here. In decoding, the forward models generate predicted patterns for each condition, and these predictions are compared with the pattern for a held-out fMRI image using the Pearson correlation. The prediction best-correlated with the held-out image is chosen as the predicted label for the held-out trial. 

An image corresponding to the proposition $\text{chase} \langle \text{cat}_a, \text{dog}_p\rangle$ is decoded as follows:\\
\textbf{Single-pattern model.} In an agent-classification trial, the agent-only model is run forward to produce the estimated pattern for each filler in the agent role, ignoring the patient. The prediction for "dog", the third filler, as a candidate agent is thus:
\begin{align*}
\begin{pmatrix} 0 & 0 & 1 & 0 \end{pmatrix} \hat{\beta}_\text{agent} = \text{fMRI}(\text{dog}_a)
\end{align*}
The filler pattern most correlated with the held-out image is chosen as the true filler. If the best-correlated filler is dog$_a$, the test trial is coded as a 1|otherwise 0. \\
\textbf{Mixed-pattern model.}: In an agent-classification trial, the patient variable is fixed to its true value|dog$_p$|while the agent variable is varied. Each agent and patient is associated with an indicator ("dummy-coded") vector whose sum is an indicator vector for the whole proposition (Appendix A):
\begin{align*}
\langle \text{man}_a, \text{dog}_p\rangle =& 
	\begin{pmatrix} 0 & 0 & 0 & 0 & 0 & 0 \end{pmatrix} \\
	&+ \begin{pmatrix} 0 & 0 & 0 & 0 & 0 & 1 \end{pmatrix} \\
	=& \begin{pmatrix} 0 & 0 & 0 & 0 & 0 & 1 \end{pmatrix}
\end{align*}
Since there is an intercept term (omitted here), the contribution of man$_a$ is reflected in the intercept coefficients corresponding to the arbitrarily chosen baseline condition $\langle \text{man}_a, \text{girl}_p\rangle$. Thus, each possible filler for the agent is modeled alongside the true patient, yielding a prediction for the entire proposition formed by placing each of the four candidate fillers into the agent role (\cref{fig_decoding}):
\begin{align*}
\begin{pmatrix} 0 & 0 & 0 & 0 & 0 & 1 \end{pmatrix} \hat{\beta}_\text{mixed} = \text{fMRI}(\langle \text{man}_a, \text{dog}_p\rangle) 
\end{align*}

\subsection{Evaluation}

Each model was evaluated using the verb-wise cross-validation procedure employed by F\&{}G. For each of the five verbs in the stimulus set, all trials with the given verb were held out for testing, and the model was fit to data from the four remaining verbs. To succeed in role-identification, the model must therefore generalize filler-role representations to new verbs. The chance level for each decoding task is 25\%. In our results, the effect sizes for individual decoding evaluations are generally small|around 1 to 2\%. This is consistent with typical results obtained in fMRI, and in particular with F\&{}G's original analysis of the same data, where the effect in both agent and patient regions did not exceed 1\%.

\section{Results} \label{sec_results}
\begin{figure*}[h!]
\begin{center}
\includegraphics[scale=.2]{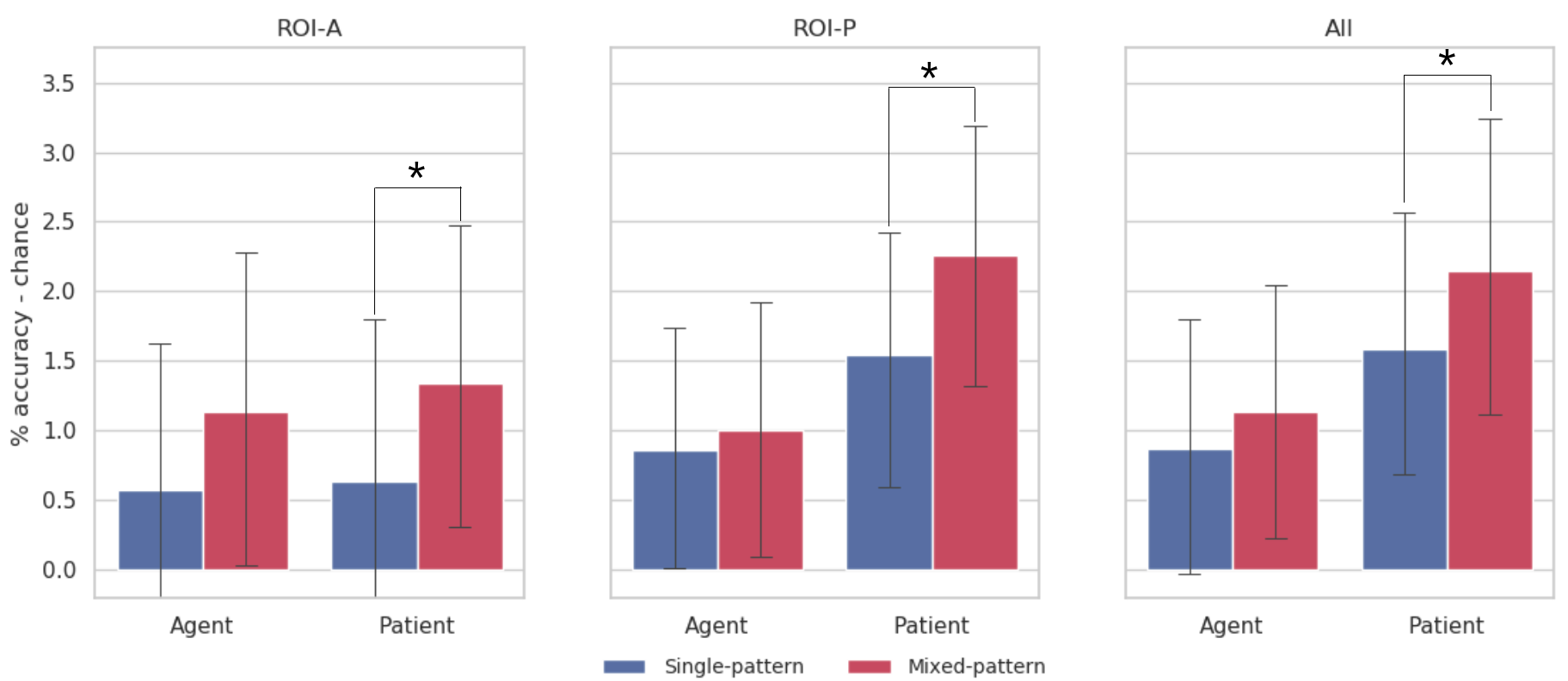}
\end{center}
\caption{\textbf{Decoding accuracy} of single- (blue) and mixed-pattern (red) models for each role and in each ROI. }\label{fig_accuracy}
\end{figure*}

The decoding results are detailed in \cref{table_accuracy} and \cref{fig_accuracy}. Since these analyses are post-hoc, we follow the convention in the original paper and report results without correcting for multiple comparisons, remarking again that the present analysis was of an exploratory nature. Significance was established using one-sample $t$-tests under the directional hypothesis that accuracy is above chance. 

\begin{table*}[h!]
\noindent
\begin{center} {\small 
\begin{tabular}{|c||c|c|c|c|c|} \hline
	&	\textbf{agent-only}	& \textbf{mixed (agent)} & \textbf{patient-only}  & \textbf{mixed (patient)}  \\ \hline\hline
\textbf{ROI-A} & .256 & .261$^*$ & .256 &  .263$^*$ \\
\hline
\textbf{ROI-P} & .259 & .260$^*$ &  .265$^{**}$ & .273$^{*{*}*}$	\\ \hline
\textbf{All} & .259$^*$ & .261$^{**}$ & .266$^{**}$ & .271$^{*{*}*}$	\\
\hline
\end{tabular} 
}		\end{center}
\caption{Decoding accuracy for each region (rows) and model (columns). \emph{mixed (agent)} and \emph{mixed (patient)} refer to results on decoding of each role using the mixed-pattern model. "All" refers to the concatenation of agent and patient ROIs, stripping any redundant voxels due to the overlap between regions. $*$: $p<.05$, $**: p<.01$, $*{*}*: p<.001$ (one-tailed t-test, uncorrected). }\label{table_accuracy}
\end{table*} 

\noindent\textbf{Single-pattern models}. Neither the agent or patient single-pattern model yielded significant classification accuracy in ROI-A, and the identification accuracies for those roles were identical in that region. In ROI-P, patient decoding was significant ($p<.01$) and agent decoding was not. Neither region by itself supported single-role decoding of agents, but the concatenation of voxels from both ROI-A and ROI-P ("All") did ($p<.05$).

\noindent\textbf{Mixed-pattern models}. Mixed-pattern decoding is significant in both regions and for both roles, as well as when using voxels from both ROIs. Accuracy is greater for the mixed-pattern models than for the single-pattern models in every case, and significantly greater between patient-only decoding and mixed-pattern decoding of agents in each region ($p=.012$ ROI-A, $p=.030$ ROI-P, $p=.047$ All).

\section{Discussion} \label{sec_discussion}

Regarding the question of orthogonal representational spaces for thematic roles, the key comparisons are between the single- and mixed-pattern models. Accuracy of decoding was higher for the mixed-pattern model in each case, and significantly higher for patients in every ROI (\cref{fig_accuracy}, ROI-A $p=.010$, ROI-P $p=.030$, All $p=.047$). In ROI-A, neither single-pattern model attains significance, but both mixed-pattern models do. These results indicate, first, that both ROI-A and ROI-P carry information about both agents and patients. Second, they indicate that, within both regions, the relevant patterns are non-orthogonal, a result inconsistent with the localist hypothesis.

Our single-role decoding results depart from the conclusions drawn in F\&{}G regarding the role-specificity of ROI-A and ROI-P based on a similar procedure. This may be due to several methodological differences. First, F\&{}G applied a linear classifier to held-out images, whereas here propositions were decoded by producing predicted patterns and comparing them to held-out images using the Pearson correlation. Second, our accuracies were obtained by predicting from the full multi-voxel patterns in each ROI, whereas F\&{}G produced an accuracy statistic for each ROI by averaging the voxel-by-voxel accuracies obtained from a searchlight centered at each voxel within the subregion. This leads to one major disanalogy between our results and theirs, which is that the sizes of the voxel populations considered in decoding differ between ROI-A and ROI-P. In particular, the set of ROI-P voxels|which differs between subjects due to the leave-one-out localization of ROIs|is generally much larger than that from ROI-A, which may explain the overall higher accuracy when using the voxels from ROI-P. 

\subsection{Role-selectivity of pattern estimates}

Distinctness of fillers from one another in ROI-A and ROI-P \ref{def_filler_discr} is verified by the above results. However, they do not speak directly to the question of whether the patterns estimated in the mixed models are themselves role-selective|i.e. whether the estimates for agents contain information about patients, and vice-versa. This is particularly important given that, in the agent region, we found no real difference between the decodeability of agents and patients. Role selectivity \ref{def_role_discr} can be evaluated by \emph{cross-decoding} \citep{allefeld&haynes2014manova}. In cross-decoding, a model is estimated using one set of conditions, and it is asked whether this model generalizes to the same condition under an additional manipulation (here, role-reassignment). 

In the present case, the model parameters are neural patterns for filler-role bindings, which can be repurposed for the decoding of fillers from the other role. To do this, the mixed-pattern model was estimated on the training set with correctly-labelled propositions. Then, in decoding, propositions in the test set were recoded to the indicator for the same proposition, but with role assignments for the two fillers in the proposition swapped. For instance, the proposition $\langle \text{man}_a,\text{dog}_p \rangle$ is recoded to $\langle \text{dog}_a, \text{man}_p \rangle$. Estimates for agents are thus used to model patients, and patient estimates model agent variables. This does not evaluate the role-selectivity of agent and representations per se, since each prediction of the mixed model contains both an agent and patient component (\cref{sec_technical_remarks}). It does, though, offer a glimpse as to the role-selectivity of these patterns.
\begin{table}[h!]
\begin{center}
{\small 
\begin{tabular}{|l||c|c|} \hline
	& \textbf{mixed (agent)}	& \textbf{mixed (patient)} \\ \hline \hline
\textbf{ROI-A}	&	.255 $p$=.124	& 	.255,	$p$=.194	\\ \hline
\textbf{ROI-P}	&  .258, $p$=.089 &	.251, $p$=.401 \\ \hline
\textbf{All} & .259$^*$ $p$=.015 & .251 $p$=.427	\\ \hline
\end{tabular}  }
\end{center}
\caption{\textbf{Cross-decoding} accuracy using mixed-pattern estimates of agents and patients from each ROI. 
} \label{table_cross_decoding}
\end{table}

\cref{table_cross_decoding} displays cross-decoding results for the mixed-pattern models estimated from each ROI. The "mixed (agent)" column indicates the accuracy of decoding agents using the patient estimates, and vice versa for the "mixed (patient)" column. While decoding of patients with agent vectors is not significant in any region, decoding of agents with patient vectors is trending in ROI-P and significant when using all voxels.

The results suggest high role-selectivity in the agent estimates, and relatively low sensitivity of patient patterns, particularly when estimated in ROI-P and over all voxels. What, then, is the character of the patient bindings in lmSTC? Possibly, these patterns include a significant common component across role assignments, reflecting a sort of "main effect" of the filler, irrespective of its role, which is especially present in ROI-P. If the proportion of signal associated with this common component is high for patients, relative to the role-selective signal, the patient patterns could be equally serviceable for the decoding of agents and patients. However, the fact that both agent and patient patterns occur in the cross-decoding predictions makes this result more difficult to interpret.

\section{Conclusion}
\label{sec_conclusion}

In \cref{sec_problem_statement}, a formally explicit model of neural computations implementing the concept of compositionality was proposed. This approach follows a hunch that reasoning about neural representations at a rather abstract level|as patterns of activation, in correspondence with cognitive symbols, manipulated and combined using simple operations|can generate detailed predictions about the representational structure of neural patterns. Furthermore, it is possible to evaluate these predictions against neural data. 

Using this approach, it was shown that modeling neural patterns for multiple constituents in complex structures, instead of just single-constituent patterns, significantly improves the classification accuracy of patients in both the "agent" and "patient" regions discovered by \citet{frankland_greene_2015}. This finding appears incompatible with a view in which patterns for distinct roles are represented over spatially disjoint populations of voxels. In both regions, patterns containing information about the assignment of semantic values to structural roles are superposed to construct patterns representing assignments of event participants to their thematic roles in a sentence. These superposed representations are not spatially partitioned, but are instead represented over a shared set of voxels. The evidence as to the role-selectivity of these estimated patterns is equivocal; however, to the extent that they are role-selective, this role-selectivity is not manifested in a localist representation of these patterns, but rather as a set of distributed filler-role bindings quite unlike the data registers of a computer.



\bibliographystyle{acl_natbib}
\bibliography{research}

\appendix

\begin{appendices}

\section{Model specifications} \label{appendix_models}

Pattern estimates were obtained from two types of models. In the \textbf{single-pattern} model, the estimates are obtained by fitting a pair of independent linear regressions containing only regressors for fillers in the agent and, separately, the patient role. The resulting model coefficients yield the mean pattern for all trials in which the given filler occurred in the given role|a single-role estimate. 

In the \textbf{mixed-pattern} model, regressors are constructed to model the \emph{mixed} data|i.e. data assumed to contain signal associated with both agent and patient roles. Since a design containing all eight regressors|one for each filler-role binding|is multicollinear, we employ the following indicator-coding scheme, chosen with the requirement that the the code for each of the sixteen possible proposition types be distinct. 
\ex. \textbf{Indicator codes for mixed-pattern regressors} \label{def_dummy_code}

\begin{align*}
\text{man}_a &= \begin{pmatrix}  0 & 0 & 0 & 0 & 0 & 0 \end{pmatrix} \\
\text{girl}_a &= \begin{pmatrix}  1 & 0 & 0 & 0 & 0 & 0\end{pmatrix} \\
\text{dog}_a &= \begin{pmatrix}  0 & 1 & 0 & 0 & 0 & 0 \end{pmatrix} \\
\text{cat}_a &= \begin{pmatrix} 0 & 0 & 1 & 0 & 0 & 0 \end{pmatrix} \\
\text{man}_p &= \begin{pmatrix}  0 & 0 & 0 & 1 & 0 & 0 \end{pmatrix} \\
\text{girl}_p &= \begin{pmatrix}  0 & 0 & 0 & 0 & 0 & 0\end{pmatrix} \\
\text{dog}_p &= \begin{pmatrix}  0 & 0 & 0 & 0 & 0 & 1 \end{pmatrix} \\
\text{cat}_p &= \begin{pmatrix}  0 & 0 & 0 & 0 & 1 & 0 \end{pmatrix} \\
\end{align*}
The resulting model coefficients are not interpretable as patterns for filler-role bindings, but instead indicate the marginal contribution of each pattern in relation to the arbitrarily chosen baseline condition $\langle\text{man}_a, \text{girl}_p\rangle$. Each setting of the indicator vector produces predicted patterns for one of the mixed conditions. For instance, the prediction for $\begin{pmatrix} 0 & 0 & 0 & 0 & 1 & 0\end{pmatrix}$ is the expected pattern for the condition $\langle \text{man}_a, \text{cat}_p \rangle$. The prediction for $\begin{pmatrix} 0 & 1 & 0 & 0 & 0 & 0 \end{pmatrix}$ is the expected pattern for $\langle \text{dog}_a, \text{girl}_p\rangle$. 

All models are fit using Generalized Least Squares (GLS), with covariance matrix along the temporal dimension estimated using the optimal Ledoit-Wolf shrinkage factor \citep{ledoit2003shrinkage}. The data are spatially pre-whitened prior to model estimation, to take into account spatial autocorrelations between voxels \citep{diedrichsen&kriegeskorte2017representational_models}. That is, model estimation included, as a prelude to fitting the predictive linear model, an estimate of the spatial correlations between voxels. Using the training data, a first-level model was fit to the unwhitened trial-wise estimates, with the first-order model using the same specification of variables as the model to be evaluated. Predictions from the first-order model were used to construct a residual series for the training data, which was then used to produce a regularized estimate of the covariance matrix|also using the Ledoit-Wolf shrinkage factor. The resulting estimate was used to render the noise in the data isotropic. Finally, the predictive model was fit to the whitened data. While the test data were whitened prior to inference, the test data were not used in the estimate of the whitening parameters. 

In preliminary analyses, univariate noise-whitening\footnote{i.e. whitening under the assumption of a diagonal covariance matrix.} was also evaluated, and led to systematically worse accuracy in classification. We conclude that spatial autocorrelations are significant in our data, requiring multivariate normalization.

\section{Dot products of mixed data} \label{appendix_mixed_dot_products}

Our decoding strategy compares predicted patterns with observed patterns by taking the dot product of the prediction with each image. This methodology is aimed at distinguishing between two cases: 
\begin{enumerate*}[label=Case \arabic{enumi}.,ref=Step \arabic{enumi}]
\item[(1)] pattern components for distinct roles are pairwise orthogonal, and \label{Case 1.}
\item[(2)] the patterns for filler-role bindings in distinct roles are non-orthogonal. \label{Case 2.}
\end{enumerate*}
Storage of filler-role bindings for distinct roles in non-overlapping regions, allocated to distinct populations of voxels, is a subcase of (1). By the same token, Case (2) implies that agent and patient representations are represented over the same population of voxels. In this Appendix, we spell out how the single-pattern and mixed-pattern models will behave in each of these cases.

Each experimental condition consists of a pair of filler-role bindings|say, $\langle\text{dog}_a,\text{cat}_p\rangle$| modeled as pattern components summed together in the neural signal to produce a pattern for each complex proposition. Each condition therefore contains a pattern corresponding to the agent, and also to the patient. For simplicity, we assume that all patterns are of approximately the same norm, so that the normalizing factors can be ignored. Let $f_{i,a}$ denote the $i$th filler in role $a$, and let $\hat{f}_{i,a}$ denote the true neural pattern for $f_{i}$ in that role. For a proposition $\langle f_{i,a}, f_{j,p}\rangle$, 
the observed image generated by the pattern is:
\ex. $\text{fMRI}(\langle f_{i,a},f_{j,p}\rangle) = \hat{f}_{i,a} + \hat{f}_{j,p} + \varepsilon$

The expected value of trials with this condition is therefore $\hat{f}_{i,a} + \hat{f}_{j,p}$. A trial is decoded by taking a normalized dot product between the prediction and the image:
\ex. ${v_\text{pred}}^\top (\hat{f}_{i,a+}\hat{f}_{j,p} + \varepsilon)$

The dot product with the highest value is then chosen as the predicted class. Since the noise is rendered isotropic by preliminary whitening, expected dot products with it are independent of the direction of the prediction, depending only on the magnitude of the pattern vector, which is normalized. Hence, we ignore dot products with the noise. 

The single-pattern model predicts $v_\text{pred} = \hat{f}_{j,a} + \frac{1}{3} \sum_{j\not = i} \hat{f}_{j,p}$ 
, where we define $ \frac{1}{3}\sum_{k\not = j} \hat{f}_{k,p} \equiv \hat{f}_{i/\cdot,p}$. The form of this mean over patient patterns is due to the omission of the diagonal conditions. 
The mixed-pattern model makes predictions of the form $v_\text{pred} = \hat{f}_{i,a} + \hat{f}_{j,p}$, where in decoding, the other role is fixed to its true value for that observation. Clearly, the dot product with the observation is achieved precisely when $v_\text{pred} =  \hat{f}_{i,a} + \hat{f}_{j,p}$, i.e. when the true value of $\hat{f}_{\cdot,a}$ is assigned. So, up to noise, the correct mixed pattern will be the best correlated with the image. 

Under spatial partitioning|case (1)|a region containing information about the agent role will contain none about the patient role, meaning that all patient patterns are the same in expectation: $\hat{f}_{j,p} = \hat{f}_{k,p}$ for any $k$. So, the single-pattern estimate likewise reaches a maximum with respect to the image. Thus, if information is spatially segregated, the single-pattern model will do just as well, and may even do better. Since the region is uninformative about the value of the other role, fewer observations are used to compute the estimate of $\hat{f}_{j,p}$, which in case (1) is in fact a single underlying pattern estimated from one-third as many observations. Hence, the estimate of $\hat{f}_{j,p} = \frac{1}{3}\sum_{k\not = j} \hat{f}_{k,p} \equiv \hat{f}_{j/\cdot,p}$ is more likely to be affected by estimation noise, reducing its reliability in decoding. On the other hand, if the patterns for different fillers are in fact distinct, then the single-role estimate $\hat{f}_{i,a} + \hat{f}_{i/\cdot,p}$ will \emph{not} attain a maximum with respect to the image. 

This by itself may not lead to changes in the relative rankings of different fillers if the patterns for all agent bindings are orthogonal to those for patient bindings. Consider an agent-decoding trial. Letting $\hat{f}_{t,a}$ denote the true agent filler, we compare the true image to the prediction for $\hat{f}_{i,a}$ in both the single-pattern and mixed-pattern cases:\\
\textbf{single-pattern}
\begin{align*}
(\hat{f}_{i,a} + \hat{f}_{i/\cdot,p} )^\top (\hat{f}_{t,a} + \hat{f}_{j,p}) &= {\hat{f}_{i,a}}^\top \hat{f}_{t,a} + {\hat{f}_{i,a}}^\top \hat{f}_{j,p} + \\
	&= {\hat{f}_{i/\cdot,p}}^\top \hat{f}_{t,a}  + {\hat{f}_{i/\cdot,p}}^\top \hat{f}_{j,p}\\
\end{align*}\\
\vspace{-1.75cm}\\
\textbf{mixed-pattern}
\begin{align*}
(\hat{f}_{i,a} + \hat{f}_{j,p} )^\top (\hat{f}_{t,a} + \hat{f}_{j,p}) &= {\hat{f}_{i,a}}^\top \hat{f}_{t,a} + {\hat{f}_{i,a}}^\top \hat{f}_{j,p} +\\
	&= {\hat{f}_{j,p}}^\top \hat{f}_{t,a} + \norm{\hat{f}_{j,p}}^2
\end{align*}
If the agent and patient fillers are pairwise orthogonal, then the two terms in the center are constant (zero). The last term ${\hat{f}_{i/\cdot,p}}^\top \hat{f}_{j,p}$ in the single-pattern decoding will not be exactly constant across choices of $\hat{f}_{i,a}$ due to differences in the mean pattern that is estimated. If many patient patterns are estimated, the differences between these estimates of the mean patient pattern will be small. Hence, we treat it as close to constant. For both models, then, the score for agent $i$ only depends on ${\hat{f}_{i,a}}^\top \hat{f}_{t,a}$ in both cases. Rankings between distinct agents are thus the same across models. By consequence, differences in their performance imply that the patterns for each role are non-orthogonal|and, by extension, spatially overlapping vectors.

\end{appendices}

\end{document}